\documentclass{article}

    \PassOptionsToPackage{numbers, compress}{natbib}

\usepackage[preprint]{neurips_2022}

\usepackage[utf8]{inputenc} 
\usepackage[T1]{fontenc}    
\usepackage{hyperref}       
\usepackage{url}            
\usepackage{booktabs}       
\usepackage{amsfonts}       
\usepackage{nicefrac}       
\usepackage{microtype}      
\usepackage{graphicx}
\usepackage{listings}
\usepackage{csquotes}
\usepackage{amsmath}
\usepackage{lipsum}
\usepackage{enumitem}
\usepackage{amssymb}
\usepackage[utf8]{inputenc}
\usepackage[linesnumbered,ruled]{algorithm2e}
\SetKwProg{Init}{init}{}{}
\usepackage[dvipsnames,svgnames]{xcolor}
\usepackage{ragged2e}
\usepackage{multirow}
\usepackage{multicol}
\usepackage{booktabs}
\usepackage{graphicx}
\usepackage{tabularx}
\usepackage[most]{tcolorbox}
\usepackage{makecell}
\usepackage{tcolorbox}
\tcbuselibrary{breakable}
\usepackage{soul}
\usepackage{pifont}
\newcommand{\cmark}{\textcolor{DarkGreen}{\ding{51}}}
\newcommand{\xmark}{\textcolor{red}{\ding{55}}}

\usepackage{subfig}

\lstdefinelanguage{Python}{
	keywords={select_text, select_shapes, select_slides, format_text, format_shapes, format_slides, insert_text, insert_shapes, insert_images, insert_slides, delete_text, delete_shapes, delete_slides},
	keywordstyle=\color{blue}\bfseries,
	ndkeywords={text, shapes, shape, slides, slide, images, image, circle, textRanges, selectedRectangles},
	ndkeywordstyle=\color{DarkCyan}\bfseries,
	otherkeywords={+, =>,<=, ==, >,< , ||},
	identifierstyle=\color{black},
	sensitive=false,
	comment=[l]{\#},
	morecomment=[s]{"""}{"""},
	commentstyle=\color{purple}\ttfamily,
	stringstyle=\color{red}\ttfamily,
	morestring=[b]',
	morestring=[b]"
}
\tcbset{
  aibox/.style={
    width=474.18663pt,
    top=10pt,
    colback=white,
    colframe=black,
    colbacktitle=black,
    enhanced,
    center,
    attach boxed title to top left={yshift=-0.1in,xshift=0.15in},
    boxed title style={boxrule=0pt,colframe=white,},
  }
}
\newtcolorbox{AIbox}[2][]{aibox,title=#2,#1}
\definecolor{aigold}{RGB}{244,210, 1} 
\definecolor{aigreen}{RGB}{210,244,211} 

\sethlcolor{aigreen}
\definecolor{aired}{RGB}{255,180,181}

\newtcbox{\mybox}[1][green]{on line,
arc=0pt,outer arc=0pt,colback=#1!10!white,colframe=#1!50!black,
boxsep=0pt,left=0pt,right=0pt,top=0pt,bottom=0pt,
boxrule=0pt,bottomrule=0pt,toprule=0pt}

\newcommand{\system}{Semantic Interpreter\xspace}
\newcommand{\dsl}{ODSL\xspace}

\title{Natural Language Commanding via Program Synthesis}

\author{
  Apurva Gandhi \\
  Microsoft \\
  \texttt{apurvagandhi@microsoft.com} \\
  \And
  Thong Q. Nguyen\\
  Microsoft \\
  \texttt{thongnguyen@microsoft.com}\\
  \And 
  Huitian Jiao \\
  Microsoft \\
  \texttt{huijia@microsoft.com} \\
  \And 
  Robert Steen \\
  Microsoft \\
  \texttt{robsteen@microsoft.com} \\
  \And 
  Ameya Bhatawdekar \\
  Microsoft \\
  \texttt{ambhataw@microsoft.com}
}

\begin{document}

\maketitle

\begin{abstract}
   We present \system, a natural language-friendly AI system for productivity software such as Microsoft Office that leverages large language models (LLMs) to execute user intent across application features. While LLMs are excellent at understanding user intent expressed as natural language, they are not sufficient for fulfilling application-specific user intent that requires more than text-to-text transformations. We therefore introduce the Office Domain Specific Language (\dsl), a concise, high-level language specialized for performing actions in and interacting with entities in Office applications. \system~leverages an Analysis-Retrieval prompt construction method with LLMs for program synthesis, translating natural language user utterances to \dsl programs that can be transpiled to application APIs and then executed. We focus our discussion primarily on a research exploration for Microsoft PowerPoint.
\end{abstract}

\section{Introduction}
Productivity apps like Microsoft Office provide a multitude of features to help users accelerate their daily work. Recently, LLMs have led to a surge of interest in building copilots or assistive experiences that allow users to interact with apps through natural language. As a core component of such experiences, a natural language commanding interface has the potential to transform every user into a power user by making an app's whole commanding surface accessible through natural language. Users can simply express what they want to do in terms of outcomes (e.g., \enquote{Bold all the keypoints} or \enquote{Insert a poem by Tagore and make it look beautiful}) and see it executed before their eyes. In this work, we explore an approach for building such a component: \system is a natural language-friendly AI system that leverages and enhances the power of large language models (LLMs) to execute user intent across application features.

To do this, \system must both \textit{understand} and \textit{fulfill} user intent expressed as natural language. LLMs excel at the former but are not sufficient for the latter. LLMs implement text-to-text transforms. Therefore, while they are excellent at understanding user intent expressed as natural language, they are fundamentally unable to fulfill application-specific user intent that requires more than text-to-text transformations (e.g., “create a new slide,” “insert a poem about hummingbirds in a blue rectangular box,” etc.)

To allow LLMs to encode application-specific user intent and communicate this to the application in a way that can be handled, we require a text-based representation of the user intent that our application is able to interpret. A \textit{program representation}, which encodes user intent with a domain-specific language (DSL), elegantly satisfies this requirement. A domain-expert can organically inject application-specific knowledge in the implementation of the DSL’s interpreter. In this way, a DSL provides a powerful abstraction that allows for a separation of concerns between intent understanding and fulfillment, letting us architect a system that appropriately leverages and fuses the strengths of LLMs with application-specific, domain-expert knowledge. 

We have seen that LLMs are capable of not just generating text but code as well \cite{chen2021evaluating, nijkamp2023codegen}. \system exploits this powerful capability of LLMs to perform \textit{program synthesis} \textendash~understanding user intent expressed as natural language and generating an executable program to fulfill this intent. \system augments the powerful but approximate capabilities of LLMs with the precision of symbolic representations by leveraging LLMs to transpile natural language user utterances to verifiable plans expressed as DSL programs. 

There are several challenges to leveraging LLMs in this way. First, LLMs are pretrained on large-scale collections of code repositories but are not fine-tuned on the APIs of a specific application. Furthermore, the resources and number of examples required for such fine-tuning may be impractical for many domains. We therefore leverage a few-shot prompting or in-context learning approach where we include relevant DSL syntax description, example user utterances, DSL programs, application context data and additional system instructions as part of the prompt. However, LLMs typically have strict token limits for context length. For example, the GPT-3 and GPT-3.5 models from OpenAI have a token length limit of 4097 tokens and while more recent models like GPT-4 have much large token limits, the cost of inference often scales with the number of tokens\footnote{\url{https://openai.com/pricing}}. To address this constraint, in addition to focusing on creating a concise DSL syntax, we use a just-in-time prompt engineering method that dynamically constructs a prompt tailored to the user utterance by choosing the most relevant DSL program examples and other prompt artifacts. We call our prompt engineering method an \textit{analysis-retrieval method} (ARM) as it builds on top of retrieval-augmented generation (RAG) \cite{NEURIPS2020_6b493230} methods with additional analysis steps that leverage classifiers to help narrow down the search space of prompt artifacts we select.

A second challenge of using LLMs with a few-shot prompting approach is that they are prone to hallucinations, often generating output that could be inaccurate or inconsistent with the prompts provided to them \cite{Ji_2023}. This is exacerbated when using the LLM to generate action plans in a DSL or API that the LLM was not pretrained on \textendash~the LLM may generate code that is semantically incorrect, have compile-time errors (syntax errors, unsupported statements/parameters, type errors, etc.) or result in runtime errors. To help reduce program synthesis errors, we create the Office Domain Specific Language (\dsl), an LLM-friendly API for performing actions in Office applications. 

Evaluation of natural language commanding systems also poses a challenge. Often, user queries are abstract and under-specified with numerous valid interpretations. For example, for a user query such as \enquote{Make the slide look beautiful,} there is no single correct interpretation. Numerous actions or combinations of them can satisfy this intent; e.g., adding an image, animating the slide, changing text formatting properties, etc. In this work, since we synthesize programs to fulfill user intent, we reformulate the problem of evaluating the natural language commanding systems into the problem of analyzing program equivalence.

In summary, our contributions are:
\begin{itemize}
    \item We discuss the design of The Office Domain Specific Language (ODSL) \textendash~a high-level, LLM-friendly language specialized for performing actions in and interacting with entities in Office Applications. 
    
    \item We describe the architecture of \system, which leverages an analysis-retrieval prompt engineering framework with LLMs to translate natural language user queries to ODSL programs that can be interpreted by Office applications. 
    
    \item We describe a procedure for evaluating natural language commanding systems that leverage program synthesis. 
\end{itemize}

While our framework is general and applies to Office applications and other productivity apps in general, we focus our discussion with an exploration for PowerPoint.

\section{Related Work}

\subsection{Program Synthesis}
Program synthesis refers to automatic generation of programs from given specifications.
There are different types of specifications that can be used for program synthesis, such as complete formal specifications \cite{10.1145/1993498.1993506, MANNA1975175, 10.1145/357084.357090, 10.1145/1168917.1168907}, input-output examples \cite{NEURIPS2021_ba3c95c2, chen2018towards, 10.1145/2240236.2240260, parisotto2017neurosymbolic, 10.1145/1707801.1706337}, and natural language descriptions \cite{10.1145/2884781.2884786}. Each type of specification has its own advantages and limitations. Complete formal specifications are precise and unambiguous, but they are difficult to write and verify by end-users. Input-output examples are easy to provide and intuitive to understand, but they may require many examples to capture the user’s intent or they may be hard to construct for some domains. Natural language descriptions, which is the focus of our work, are natural and expressive, but they may be incomplete or ambiguous. 

There are several existing works that use natural language descriptions for program synthesis in different domains. For example, PAL \cite{gao2023pal} and PoT \cite{chen2022program}  generate Python programs which are offloaded to an external interpreter as the intermediate reasoning steps to solve problems expressed in natural language. Codex \cite{chen2021evaluating}, CodeGen \cite{nijkamp2023codegen} and PaLM \cite{chowdhery2022palm} are LLMs that can generating code from natural language input description. There are also earlier approaches that construct novel neural-network architectures optimized for program synthesis \cite{das2022sketchode, ling-etal-2016-latent, yin-neubig-2017-syntactic, iyer-etal-2018-mapping}. Synthesizing programs in general-purpose programming languages, however, has shown several challenges and limitations in essential aspects such as handling inconsistency in natural language, generating correct and efficient programs, and ensuring semantic grounding and alignment between natural language and code \cite{austin2021program}. There are also safety concerns due to hallucination risks associated with LLMs \cite{Ji_2023}. \system aims to address these issues by grounding program synthesis in relevant program samples selected using our analysis-retrieval method for prompt construction. Furthermore, we combine LLM-based program synthesis with static program analysis to find and auto-correct common errors in the generated program, taking a neurosymbolic approach to improve robustness of program synthesis.

\subsection{Neurosymbolic Methods}
Neurosymbolic methods integrate deep learning with symbolic representations \cite{PGL-049}. The objective is to learn functions from data, as in classical machine learning. However, unlike conventional machine learning methods that learn black-box models, neurosymbolic programming aims to learn interpretable and verifiable symbolic representations like programs in a domain-specific language (DSL) that are consistent with the given data. Neurosymbolic programming offers several advantages over traditional machine learning approaches, such as data efficiency, generalization, and explainability \cite{Hamilton_2022, susskind2021neurosymbolic}.

Several works have explored the use of neurosymbolic programming in various applications, such as program induction \cite{parisotto2017neurosymbolic}, causal effect estimation \cite{reddy2023nester}, computer graphics \cite{10.1111:cgf.14775}, task-oriented dialog systems \cite{Andreas_2020}, and scientific discovery acceleration \cite{sun2022neurosymbolic}. 
One of the main challenges in neurosymbolic programming is to design a DSL that is expressive enough to capture the desired functionality, but also suitable to enable efficient and accurate program synthesis. Moreover, the DSL should be aligned with the inductive biases and heuristics that are relevant for the task domain. For example, in the domain of regular expression based string transformations, Parisotto et al \cite{parisotto2017neurosymbolic} designed a DSL that incorporates common string operations and regular expression syntax. In the domain of causal inference, Reddy and Balasubramanian \cite{reddy2023nester} designed a DSL that encodes assumptions and constraints from causal inference literature. In this work, we introduce a DSL that captures the functionality and semantics of Office commanding. Additionally, \system includes a syntax validation and code correction procedure to ensure robustness of the synthesized program.

\subsection{LLMs for Tool Learning}
The development of LLMs that exhibit reasoning-like capabilities \cite{bubeck2023sparks} has inspired a lot of research on using LLMs for decision making tasks. Chain-of-Thought (CoT) is among pioneering works that demonstrate how LLMs can emulate a \enquote{thinking procedure} for solving problems \cite{wei2023chainofthought}. Several extensions and variations of CoT have been proposed, such as least-to-most prompting \cite{zhou2023leasttomost}, zero-shot CoT \cite{kojima2022large}, self-consistent reasoning \cite{wang2023selfconsistency}, self-ask prompting \cite{press2022measuring}, RCI framework \cite{kim2023language}, and tree of thoughts \cite{yao2023tree}.  

These works demonstrate that LLMs can mimic reasoning and decision making via natural language. However, natural language texts may not be sufficient or efficient for solving some problems that require complex calculations, translations, or information retrieval. Therefore, a new paradigm has emerged for tool learning with LLMs, which leverages external tools through text-based DSL statements or API calls \cite{qin2023tool}. For example, Cobbe et al. \cite{cobbe2021training} demonstrate how an LLM can leverage a calculator to perform basic arithmetic operations in natural language. Parisi et al. \cite{parisi2022talm} propose a TALM framework for interleaving text-based API calls (such as a QA system and a calculator) with the natural language output of LLMs. They also present a method for iteratively bootstrapping tool-use examples to improve the LLM’s proficiency in using a tool. Schick et al. \cite{schick2023toolformer} generalize TALM's approach to a wider range of simple tools, including a machine translation system, a Wikipedia search engine, and a calendar, and introduce Toolformer, an LLM that can seamlessly interact with these tools. 

Tool learning has been applied in various real-world scenarios. For example, WebGPT \cite{nakano2022webgpt} fine-tunes GPT-3 \cite{NEURIPS2020_1457c0d6} to interact with Bing search engine and outperforms human experts in information retrieval. WebShop \cite{yao2022webshop} creates a web-based environment where an agent can browse and purchase products based on human instructions. Visual ChatGPT \cite{wu2023visual} interleaves various vision foundation models with ChatGPT to enable understanding and generating images. HuggingGPT \cite{shen2023hugginggpt} connects existing models hosted by HuggingFace using a universal language interface, where the LLM serves as the orchestrator for task planning and calls existing models to handle tasks in specific domains, such as object detection and question answering. To our knowledge, \system is the first application in the Office productivity space that leverages tool-oriented learning with LLMs. 

\section{Approach}

\begin{figure}[h!]
\includegraphics[width=\linewidth]{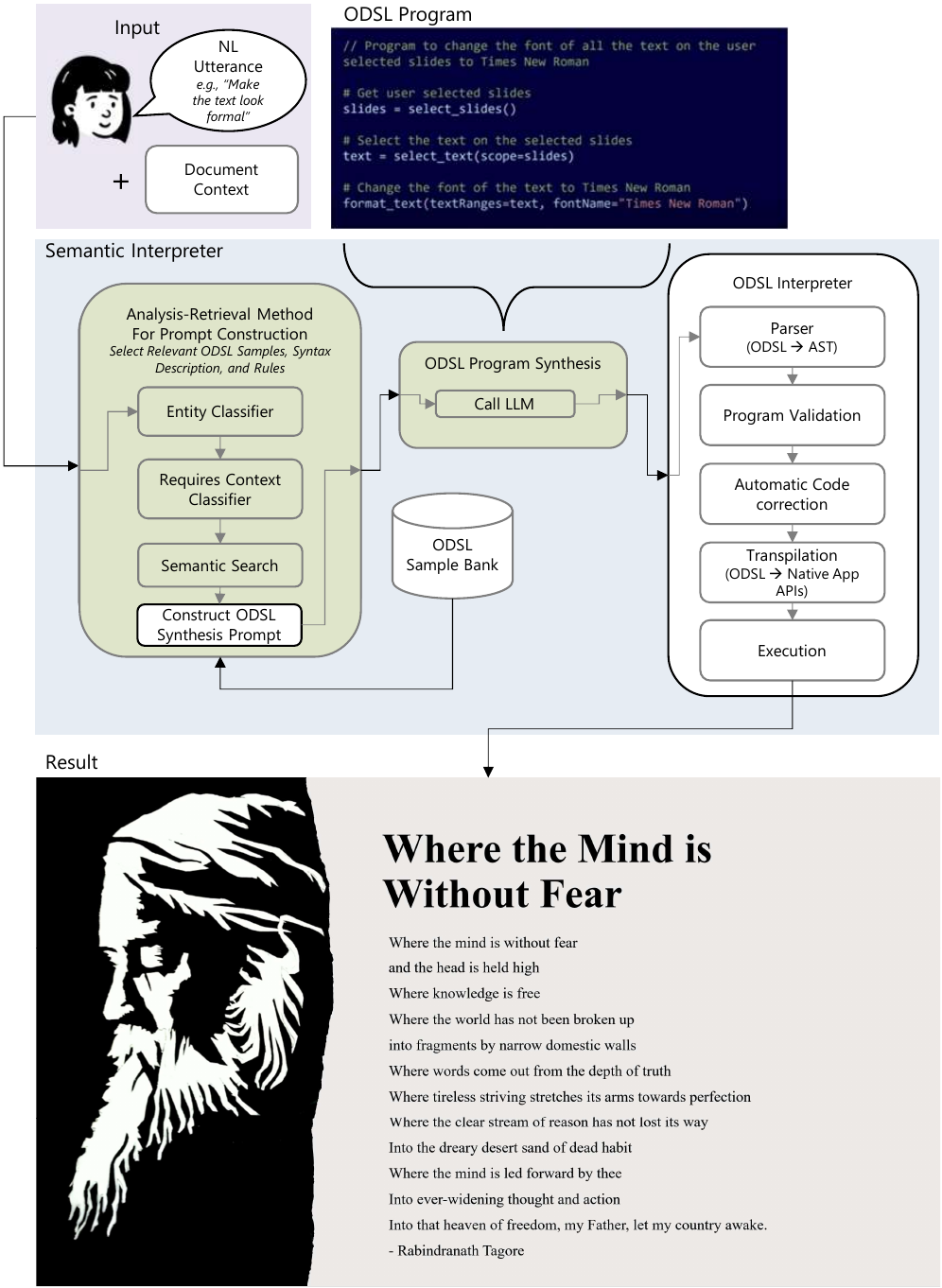}
\caption{Illustration of \system's architecture and overall approach. \system translates a user utterance to an \dsl program by leveraging an an analysis-retrieval few-shot prompting approach with an LLM for program synthesis. The \dsl program is then validated and transpiled to the app's native APIs for execution.}
\label{fig:birdeye}
\end{figure}

Fig.~\ref{fig:birdeye} illustrates the overall approach. \system takes as input a natural language user utterance and relevant document context. The document context is encoded as a light-weight JSON tree data structure that includes document entities, content and properties that are relevant to the user utterance; Fig~\ref{fig:context_aware_design} provides an example. \system then uses an analysis-retrieval method (ARM) to generate a prompt conditioned on the user utterance and document context. We discuss the ARM prompt generation method in more detail in Section~\ref{sec:system}. This prompt is fed into an LLM to generate a program representation of the user utterance in the Office Domain Specific Language (\dsl) \textendash~ an LLM-friendly DSL for performing actions in and interacting with content in Office applications. Detailed description and design principles of \dsl are laid out in Section~\ref{sec:odsl}. The \dsl interpreter parses the \dsl program into an Abstract Syntax Tree (AST) and performs analysis to validate the program and identify errors. The AST is then passed through a code-correction layer that leverages domain-specific knowledge to automatically fix bugs in the program when can be done unambiguously or using some opinionated rules. Finally, the interpreter transpiles the \dsl AST to a program written in native application APIs (e.g., Office-JS\footnote{\url{https://learn.microsoft.com/en-us/office/dev/add-ins/reference/javascript-api-for-office}}) which is then executed by the application to fulfill the user intent.

\section{The Office Domain Specific Language (\dsl)}
\label{sec:odsl}
The Office Domain Specific Language (\dsl) is a high-level commanding language for representing actions to perform in Office Apps. The language is designed with application-specific syntax that encodes the commanding surface for each app. We can think of ODSL as an LLM-friendly API for Office Applications.

Today, while applications often have existing commanding APIs as libraries as part of general-purpose programming languages (JavaScript, C++, etc.), using these libraries as targets for LLM-based program synthesis leads to suboptimal performance in practice. 

General-purpose programming languages are challenging targets for program synthesis as by nature they are not scoped in design. For example, while LLMs often tend to output a basic program that seems reasonable, on closer inspection we see that they tend to hallucinate or forget details like import statements, properties on objects, library function names, etc.

Using a general-purpose language also makes it challenging to ensure safe code. Unlike a scoped DSL, using a general-purpose language encourages the model to use all available language features and libraries, some of which are potentially unsafe.

Furthermore, general-purpose languages often have a multitude of ways to do the same thing. Beyond making the language difficult for the model to learn in a few-shot manner, this also reduces and/or complicates our ability to perform domain-specific code analyses and inferences. As we discuss in Section~\ref{sec:auto-correct}, our constrained language design in ODSL allows us to easily maintain an automatic code correction layer that can be used to auto-correct buggy code generated by LLMs in many situations. 

While in this paper we explore creating an \dsl specification for PowerPoint, other apps can build specifications that adhere to a similar design philosophy that consists of three fundamental objectives: (1) Expressive and Extensible, (2) ML-Aware and (3) Robust.

\subsection{Expressive and Extensible}\label{sec:expressive} The DSL should make it easy to express a wide variety of user intents, both descriptive and abstract, not only today but also in the future as the app's functionality is extended. The building blocks of \dsl are \textit{entities} that represent the core data structures (and associated hierarchy) in the app and \textit{statements} that perform operations on these entities. \dsl is easy to extend with new functionality by simply adding a new entity and/or statement type.

\subsubsection{Entities}
Entities are the main application-specific data structures that represent the types of objects that a user would want to create and manipulate in the application. Entities can have a has-a relationship with other entities defining a hierarchy. For example, in PowerPoint, the entity hierarchy is \texttt{slide} $\rightarrow$ \texttt{shape} $\rightarrow$ \texttt{textRange}, where a slide can contain one or more shapes and a shape can optionally contain a textRange. To define this entity hierarchy, we closely follow the data structure and hierarchy defined in the PowerPoint's open-source Office-JS and OOXML\footnote{\url{https://learn.microsoft.com/en-us/office/open-xml}} specifications. These specifications use shape as an umbrella entity to represent many interesting data types such as geometric shapes, lines/curves, charts, textboxes, etc. TextRanges are contiguous ranges of text and therefore contain smaller textRanges within them.

\subsubsection{Statements}
Statements in \dsl are used to perform operations to interact with and create new entities in Office applications. Statements use syntax that loosely resembles function calls in Python, but unlike Python have the convention of using \texttt{snake\_case} only for statement names and \texttt{camelCase} for parameter and variable names. For each entity type, we have the following four types of statements: 

\paragraph{Select Statements} These statements get a reference to an entity in the scope provided. We can provide conditions to filter the selection on using additional statement parameters.
\begin{lstlisting}      [aboveskip=\smallskipamount,belowskip=\smallskipamount,caption=Examples of select statements., label={lst:select-examples}]
# Gets all textRanges matching the string "Hello" from provided shapes.
textRanges = select_text(scope=shapes, text="Hello")

# Gets the second triangle in the Current selection.
shape = select_shapes(shapeType="Triangle", index=1)
\end{lstlisting}

    All \texttt{select} statements have a \texttt{scope} parameter to specify the scope from which to select an entity. An entity can be selected from any other entity above it in the hierarchy but not below it. For example, you can provide a scope of type \texttt{shapes} to a \texttt{select\_text} statement, but it is illegal to provide a scope of type \texttt{text} to a \texttt{select\_slides} as slides are above textRanges in the entity hierarchy and therefore textRanges cannot contain slides within them. Additionally, the \texttt{scope} parameter can be provided with the special arguments \texttt{"Presentation"} and \texttt{"Selection"} which correspond to the whole presentation and current user selection respectively. If no scope parameter is used, it defaults to the user selection.

    The rest of the parameters in the select statement act as filter conditions to give more fine-grained control on the entities to select. Select statements provide a declarative abstraction for iterating over the document object model and obtaining references that satisfy filter criterion. 
    
\paragraph{Insert Statements} These statements create new instances of entities using the provided parameters and return a reference to them.

\begin{lstlisting}      [aboveskip=\smallskipamount,belowskip=\smallskipamount,caption=Examples of insert statements., label={lst:insert-examples}]
# Inserts new "Title and Content" slides after provided ones.
slides = insert_slides(precededBy=slides, layout="Title and Content")

# Insert a textbox into each of the slides provided.
textbox = insert_shapes(shapeType="Textbox", slides=slides)
\end{lstlisting}

\paragraph{Format Statements} These statements update (usually formatting) properties of the entities. They contain many optional parameters that can be used to specify how to format the entities passed to them.

\begin{lstlisting}      [aboveskip=\smallskipamount,belowskip=\smallskipamount,caption=Examples of format statements., label={lst:format-examples}]
# Applies a set of formatting updates to provided shapes.
format_shapes(shapes=shapes, fillColor="teal", fillTransparency=0.2, top=50, left=50, height=300, width=200, lineColor="#964B00", lineTransparency=0.2)

# Formats the text in textRanges with a set of formatting properties. 
format_text(textRanges=textRanges, bold=true, fontName="Times New Roman", horizontalAlignment="Left", color="teal", italic=true, underline="Wavy")
\end{lstlisting}

\paragraph{Delete Statements.} These statements delete all the instances of entities passed to the statement. 
\begin{lstlisting}      [aboveskip=\smallskipamount,belowskip=\smallskipamount,caption=Examples of delete statements., label={lst:delete-examples}]
# Deletes shapes provided to the shapes parameter.
delete_shapes(shapes=shapes)

# Deletes the text in the textRanges
delete_text(textRanges=textRanges)
\end{lstlisting}

Notice parallels between this model and the simple but highly expressive CRUD (create, read, update and delete) model used in REST APIs, SQL, etc. We found that these statements are expressive enough to represent the majority of user utterances, while also keeping the language constructs compact and limited.

In addition to the entity-specific statements above, \dsl also allows for \textit{delegation statements} that are wrappers around specialized services or models (e.g., text-to-image model, spell-checker, etc.) This allows Semantic Interpreter to automatically extend its capabilities with specialized intelligence capabilities seamlessly and with little effort. An example of a delegation statement in \dsl for PowerPoint is an \texttt{insert\_images} statement that can internally delegate to models such as OpenAI's DALL$\cdot$E 2 \cite{ramesh2022hierarchical}.

\begin{lstlisting}      [aboveskip=\smallskipamount,belowskip=\smallskipamount,caption=Examples of \texttt{insert\_images} delegation statement., label={lst:delegation-examples}]
# Insert images consistent with description into slides provided using DALLE-2.
images = insert_images(slides=slides, description="A man walking a dog.")
\end{lstlisting}

\subsection{ML-Aware} Designing a custom DSL gives us the unique opportunity to design a language that is conducive to program synthesis via LLMs. We discuss some of our design choices to help make the \dsl LLM-friendly.

\subsubsection{Uniformity} LLMs like GPT models are trained to take a string prompt and generate new text that follows and is consistent with that prompt. To do this, LLMs must recognize patterns in the prompt and generate new text that are consistent with those patterns. As we discuss in Section~\ref{sec:system}, we provide LLMs with a prompt that contains program examples written in \dsl. The more uniform or formulaic the language is, the easier it is for the LLM to pick up on these patterns in a few-shot setting and generate new programs that are valid and consistent with the language's grammar.

As discussed in Section~\ref{sec:expressive}, \dsl has the same four types of statements across each entity and each statement follows the same naming conventions as well. Similarly, each type of statement follows similar conventions across entities; e.g., all \texttt{select} statements have a \texttt{scope} parameter followed by additional parameters that act as filter conditions for selection as discussed above. Even delegation statements like \texttt{insert\_images} follow the same naming conventions as the other statements to help the model generalize. Uniform statement types across entities also lead to common access and manipulation patterns; e.g., every \texttt{format} or \texttt{delete} statement acting on a particular entity variable is always preceded by a corresponding \texttt{insert} or \texttt{select} statement. Another choice made was to make no distinction between lists of entities and singular entities: \texttt{select} and \texttt{insert} statements always return a list of zero or more entities and similarly \texttt{format} and \texttt{delete} statements operate on a list of zero or more entities.

\subsubsection{Compactness}
A major constraint in using LLMs is the token limit for prompt length. For example the GPT-3 models have a limit of 4097 tokens for inference, in which we have to fit both the prompt and the model output. While newer models now have much more liberal token limits, the cost of inference generally scales proportional to the number of tokens.

As we discuss in Section~\ref{sec:system}, the prompt we provide to the LLM in our approach for program synthesis is quite crowded, including \dsl syntax description, few-shot \dsl program examples, document context examples and the current user utterance and document context. To fit all of this in a limited token budget, it is critical to design a DSL that has compact syntax to reduce the length of the syntax description and length of the program representations both in examples we provide to the LLM and the LLM response. 

We have found that using the four statement types per entity is a good abstraction, keeping the language expressive enough to represent most user intents while still being minimal and not having too many statements or constructs.

To see the importance of choosing appropriate statement design for compactness, consider the example in Listing~\ref{lst:compactness-comparison} which compares \dsl to another possible approach where instead of having a single statement to handle formatting like we do in \dsl, we decide to create a separate statement for each formatting property. Compared to \dsl, this latter approach is much less compact and consumes more than 1.5x the number of tokens with no functional advantage. 

\begin{lstlisting}[aboveskip=\smallskipamount,belowskip=\smallskipamount,caption=Importance of compact DSL design., label={lst:compactness-comparison}]
# ODSL example of formatting some text with various properties
# This program consumes 50 tokens with the GPT-3 tokenizer.
text = select_text()
format_text(textRanges=text, bold=true, fontName="Times New Roman", horizontalAlignment="Left", color="teal", italic=true, underline="Wavy")
-----------------------------------------------------------------------------
# Alternate possible DSL with separate statement for each formatting property.
# These programs consumes 78 tokens with the GPT-3 tokenizer.
text = select_text()
set_bold(text, true)
set_font_name(text, "Times New Roman")
set_horizontal_alignment(text, "Left")
set_font_color(text, "teal")
set_italic(text, true)
set_underline(text, "Wavy") 
\end{lstlisting}

\subsubsection{Minimizing Redundancy} 
While it is not uncommon in traditional API design to build in multiple ways to do the same thing, out of convenience for programmers, this is suboptimal when designing languages/APIs for program synthesis with LLMs. With multiple ways to do the same thing, we can confuse the model and it is harder for the LLM to learn and generalize on how to perform a particular task from a few examples when the examples themselves are not consistent in how they perform the task.

One example of a choice made to reduce redundancy in \dsl is not repeating formatting properties in \texttt{insert} statements. For example, while we could include formatting properties like \texttt{bold, fontName}, etc. in the \texttt{insert\_text} statement, we do not as one can achieve the same functionality by chaining the \texttt{insert\_text} statement with a \texttt{format\_text} statement to update these properties.

\subsubsection{Document Context Co-design}
\begin{figure}[ht!]
\centering
\includegraphics[width=\textwidth]{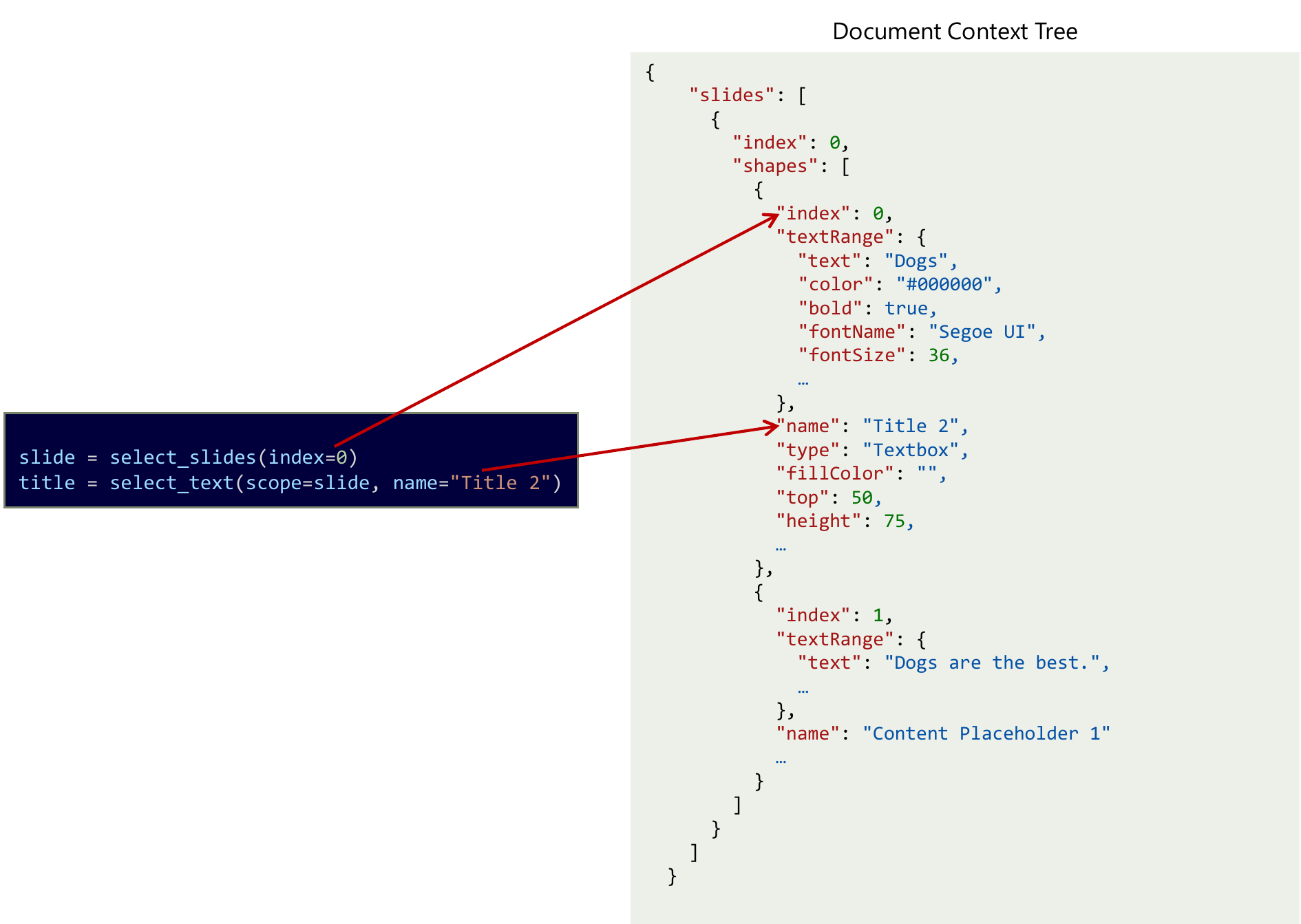}
\caption{Illustration of how \dsl statements are aware of the document context tree included in the \dsl synthesis prompt. Here we see that \texttt{select} statements have parameters that match keys of entity identifiers in the context tree.}
\label{fig:context_aware_design}
\end{figure}

To design the interaction model of how a program can represent referencing and manipulating existing entities in the application, we need the syntax to be context-aware – knowing what information the model has when synthesizing the program. Particularly, when designing the document context tree, we include entity identifier metadata (e.g., index and names) with keys that match the \dsl select statement parameter names, making it easier for a model to describe the entities it wants references to (shown in Fig.~\ref{fig:context_aware_design}). 

\subsection{Robust} 
\label{subsec:robust}
Given that the DSL is the interface between the LLM and the application, the design of the DSL has direct influence on the robustness of the overall system. Below we discuss a few strategies when designing ODSL and its interpreter to improve the overall robustness of the system.

\subsubsection{Constraining the DSL to Safe Operations} We are intentional about the scope of the language, including only statements that cannot take us to an irreversible or invalid statements. Particulary, we limit ourselves to content generation and manipulation operations that can be reversed by the user with a simple undo operation. We do not include any operations that can take us to an illegal state, such as a \texttt{close\_file()} statement.

\subsubsection{Syntax Validation for Compile-time Errors and Minimizing Runtime Errors} Whenever possible, we try to catch errors at compile time rather than runtime. Runtime errors can lead to suboptimal user experience where we halt a program in the middle of execution. Especially for abstract user intents like \enquote{make the slide beautiful,} this can lead to confusing user experience. For example, an image could be inserted on top of existing content, but the program may abort before repositioning the image to a free spot on the slide, making the resulting slide ironically look uglier than before any action was taken. 

Compile-time errors are cleaner as they do not lead to partial program execution. If the system encounters a compile time error, it can surface a more informative error to the user, rather than executing a few statements and then failing unexpectedly. \dsl is statically typed, allowing us to perform robust compile-time syntax validation. Types span both the entity types that can be returned by statements, and also fine-grained literal types which include the standard primitive types like \texttt{Number}, \texttt{String}, \texttt{Boolean} to more fine-grained and complex types like \texttt{RangedNumber}, \texttt{StringOrRegex}, etc. Examples of errors that syntax validation catches include type errors, variable out-of-scope errors, parsing errors, undefined statements, etc.

Of course, not all errors can be caught at compile time. Still, common types of runtime errors are designed to execute as no-ops (e.g., selecting and manipulating an entity that does not exist) but not a crash. 

\subsubsection{Automatic Code Correction}\label{sec:auto-correct} An advantage of having an intentionally scoped, custom DSL is that we can use application-specific knowledge to auto-correct buggy code or at least use opinionated strategies to recover from certain types of errors. This greatly helps improve accuracy of code generation when working with LLMs. The \dsl interpreter consists of a code correction layer as a mechanism to auto-correct many compile-time errors we encounter. 

\begin{lstlisting}[aboveskip=\smallskipamount,belowskip=\smallskipamount,caption=Examples of ODSL Automatic Code Correction., label={lst:correction-examples}]
# Example 1: Fuzzy Enum Matching - Invalid Program: Typo in fontName
text = select_text()
format_text(textRanges=text, fontName="Cmic Sans")

# Auto-corrected program: Cmic Sans -> Comic Sans Ms
text = select_text()
format_text(textRanges=text, fontName="Comic Sans MS")
-----------------------------------------------------------------------------
# Example 2: Statement Aliasing - Invalid Program: Incorrect statement name.
slide = select_slides() 
insert_picture(slides=slide, description="A picture of a cat") 

# Auto-corrected program: insert_picture -> insert_images 
slide = select_slides() 
insert_images(slides=slide, description="A picture of a cat") 
-----------------------------------------------------------------------------
# Example 3: Statement Injections - Invalid Program: Invalid enum value (Circle)
slide = select_slides() 
insert_shapes(slides=slide, shapeType="Circle") 

# Auto-corrected program: Circle -> Ellipse & injected format_shapes 
slide = select_slides() 
circle = insert_shapes(slides=slide, shapeType="Ellipse") 
format_shapes(shapes=circle, height=100, width=100) 
-----------------------------------------------------------------------------
# Example 4: Argument Value Correction - Invalid Program: Out-of-range value
selectedRectangles = select_shapes(shapeType="Rectangle") 
format_shapes(shapes=selectedRectangles, fillTransparency=100) 

# Auto-corrected program: fillTransparency clamped to [0, 1] range 
selectedRectangles = select_shapes(shapeType="Rectangle") 
format_shapes(shapes=selectedRectangles, fillTransparency=1) 
\end{lstlisting}

\section{Analysis-Retrieval Method for ODSL Program Synthesis}
\label{sec:system}

In this section, we describe our approach to use an LLM to generate \dsl programs that satisfy user intent by leveraging an \textit{analysis-retrieval method} (ARM) for prompt construction. 

A major constraint in many LLM systems is the token length limit. For example, the GPT-3.5 model family from OpenAI has a token length limit of 4097 tokens per API call, which in our approach, consists of the various prompt components listed in Table \ref{tab:prompt}. ARM is therefore designed to optimize for \dsl program synthesis under a constrained number of allowable prompt tokens. This method consists of two steps: (i) \textit{Analysis}, where the user utterance is matched to a set of relevant ODSL entities used to filter ODSL syntax and other prompt artifacts shown to the LLM in the prompt, and (ii) \textit{Retrieval}, where we use the analysis results to enhance the semantic search process to choose sample utterances and ODSL program examples similar to the user utterance from our ODSL sample database.

\begin{table}[ht]
\caption{Prompt components to generate a \dsl program}
\label{tab:prompt}
\setlength\extrarowheight{3pt}
\begin{tabularx}{\linewidth}{ l X }

 \hline\hline
 Components & Description \\ [0.5ex] 
 \hline\hline
 System instruction & General introduction about Semantic Interpreter and \dsl.  \\ 
 \hline
 \dsl syntax* & Syntax examples of \dsl statements for entities associated with the user utterance. \\
\hline
 Rules* & Additional guidelines and instructions associated with entities to help condition the LLM further. \\
 \hline
 Few-shot \dsl samples* & Several examples of user utterances, document context (optional), and corresponding \dsl programs selected via semantic search\\
 \hline
 Input utterance & Utterance from user. \\
 \hline
 Current document context* & Document context tree encodes document context and helps ground generation in current application context. Included optionally based on result of the Requires Context classifier.\\
 \hline\hline
\end{tabularx}
\begin{flushright}
* dynamically selected by ARM
\end{flushright}
\end{table}

Specifically, given a user utterance $x$ and document context $c$, ARM's analysis step includes a classifier $p_\alpha(e|x, c)$ to determine a set $\mathcal{E}$ of ODSL entities $e$ associated with $x$ and whether the context $c$ is needed for program synthesis. Detailed implementation of $p_\alpha(e|x, c)$ is described in Section \ref{subsec:analysis}. 

The retrieval step then leverages the analysis results to obtain a set $\mathcal{Z}$ of relevant \dsl sample programs $z$ from our database using a process called \textit{entity-aware semantic search} $p_\rho(z|x,c,e)$. This analysis-retrieval process can be represented as follows:

\begin{equation}
p(z|x, c) = \sum_{e\in\mathcal{E}} p_\rho(z|x,c,e)p_\alpha(e|x, c)
\end{equation}

We then use the retrieved sample set $\mathcal{Z}$ to condition the LLM to synthesize the \dsl program output $y$, similar to retrieval-augmented generation \cite{NEURIPS2020_6b493230}:
\begin{equation}
p(y|x,c) = \sum_{z\in \mathcal{Z}}p(z|x, c) p(y|x, c, z)
\end{equation}

\subsection{Analysis: Associated Entities and Document Context}
\label{subsec:analysis}

\begin{figure}
\begin{AIbox}{Entity \& Requires-Context Classifier}
{\bf System Instruction:}
There are 5 categories of entities in a PowerPoint presentation: text, image, shape, slide, presentation. You need to perform the following tasks:\\

1. Categorize a given sentence into entity categories. Each sentence can have more than one category.\\
2. Classify whether a sentence requires context. Context is required when additional information about the content of a presentation is required to fulfill the task described in the sentence. \\
- Adding an image about a given topic does not require context. \\
- Adding new text needs context to decide where to place the text on the current slide. \\\\
...\\

Let's think step by step.
Here are some examples:\\

{\bf User}: Make the title text on this slide red\\
{\bf Assistant}:\\
Categories: text\\
Thoughts: We can select the title text and make it red without knowing the existing text properties. Therefore we do not need context.\\
RequiresContext: false\\

{\bf User}: Add text that's a poem about the life of a high school student with emojis.\\
{\bf Assistant}:\\
Categories: text\\
Thoughts: We need to know whether there is existing text on the slide to add the new poem. Therefore we need context.\\
RequiresContext: true\\

...\\
  
{\bf User}: Create a presentation about climate change.\\ 
{\bf Assistant}:

\end{AIbox}
\caption{Sample prompt for entity and context classification.}
\label{fig:entityclassifier}
\end{figure}

The analysis component $p_\alpha(e|x,c)$ consists of 2 steps: \dsl entity classification and context classification. The entity classifier associates a given utterance with a set of entities. Note that these entities can be a superset of the entities described in \dsl, to allow for more fine-grained analysis. For example, in PowerPoint, although \texttt{images} are a subset of the umbrella \texttt{shapes} entity, we separate them here for more fine-grained selection of prompt artifacts. The entities that we classify the user utterance into for PowerPoint are \texttt{presentation, slide, text, image,} and \texttt{shape}. The classified entities are used as input along with the user utterance to a requires-context classifier to determine whether the document context $c$ is needed to fulfill the intent.

There are many ways to implement the entity and requires-context classifiers. In our experiments, we leverage LLMs to implement these with few-shot prompting. We use a single LLM call to jointly perform both classifier tasks, leveraging CoT technique to improve classification performance \cite{wei2023chainofthought}. An illustrative example of the prompt format for the entity and requires-context classifiers is presented in Fig. \ref{fig:entityclassifier}.

\subsection{Retrieval: Entity-Aware Semantic Search}
The retrieval component $p_\rho(z|x,c,e)$ based on dense passage retrieval (DPR) \cite{karpukhin-etal-2020-dense} retrieves relevant samples similar to the input utterance for few-shot prompting. However, unlike DPR which chooses samples purely based on semantic similarities, ARM's retrieval component ensures the selected samples represent all entities associated with the user utterance. This process includes 5 steps:

\paragraph{Sample Bank Construction}
We construct a sample bank $\mathcal{B}$ that consists of pairs of sample utterances and the corresponding \dsl programs. Each sample can also include sample document context $c$ if needed, and is tagged with a set of associated entities used by the analysis process to filter examples based on matched entities.

\paragraph{Sample Utterance Normalization}
To help prevent spurious matches when performing a semantic search over sample utterances in $\mathcal{B}$, we standardize the sample utterances in a process called \textit{normalization}. In this process, we modify the sample utterance to remove specificity and standardize all intent-descriptive verbs. For examples, all generative actions, such as \texttt{add}, \texttt{insert}, \texttt{create}, \texttt{generate}, are standardized as \texttt{add}; all descriptive keywords are changed into generic descriptions, so that \enquote{red} and \enquote{blue} become \enquote{color,} \enquote{funny style} or \enquote{serious tone} becomes \enquote{a given style,} etc.

\paragraph{Sample Utterance Embedding}
The embedding process is similar to DPR. We use an LLM-based encoder $\mathbf{E}$ to embed each normalized utterance $\bar{b}_i \in \mathcal{B}$ into an embedding vector $q_i = \mathbf{E}(\bar{b}_i)$. During runtime, the same encoder is used to compute the embedding vector $\mathbf{E}(x)$ for the input utterance $x$.

\paragraph{Entity-Aware Dynamic selection}\label{subsec:retrieval}

\begin{algorithm}
    \caption{The entity-aware dynamic selection algorithm}
    \label{algo:selection}
    \SetKwInOut{Input}{Input}
    \SetKwInOut{Output}{Output}

    \underline{function EntityAwareDynamicSelection} $(x,\mathcal{B},\mathcal{E}, k) \rightarrow \mathbf{R}$\;
    \Input{User utterance $x$, sample bank $\mathcal{B}$, list of associated \dsl entities $\mathcal{E}$, and minimum number of samples to retrieve $k$}
    \Output{A set of retrieved samples $\mathbf{R}$}

    Relevant samples $\bar{\mathcal{B}} \leftarrow \{\}$\;
    List of samples and similarity scores pairs $\mathcal{S}\leftarrow [ ]$\;
    \ForEach {$b_i \in \mathcal{B}$}
    {
        \If {entities$(b_i) \subset \mathcal{E}$} 
        {
            $\bar{\mathcal{B}}$.insert$(\bar{b}_i)$\;
        }
    }
    
    \ForEach {$\bar{b}_i \in \mathcal{\bar{B}}$}
    {
        $s_i \leftarrow \mathbf{E}(x)^\top \mathbf{E}(\bar{b}_i)$\;
        $\mathcal{S}$.append([$s_i, b_i$])\;
    }
    $\mathcal{S} \leftarrow \mathcal{S}.\textrm{sortBy}(s_i)$\;
    Top matches by similarities $\alpha \leftarrow \mathcal{S}[0:k]$\;
    Top matches by entities $\beta \leftarrow []$\;
    \While {length($\beta$) $\leqslant$ length$(\mathcal{E})$} 
    {
        \ForEach {$\{s_i,b_i\} \in \mathcal{S}$}
        {
            \If {entities$(b_i) \not\subset$ entities$(\beta)$}
            {
                $\beta$.append([$s_i, b_i$])\;
            }
        }
    }
    $\mathbf{R} \leftarrow \beta.\textrm{concat}(\alpha).\textrm{removeDuplicates}()$\;
    $\mathbf{R} \leftarrow \mathbf{R}[0:\max{(k, \textrm{length}(\mathcal{E}))}].\textrm{sortBy}(s_i, \textrm{desc})$\;
    \Return $\mathbf{R}$
\end{algorithm}

\begin{figure}
     \centering
     \includegraphics[width=1.0\linewidth]{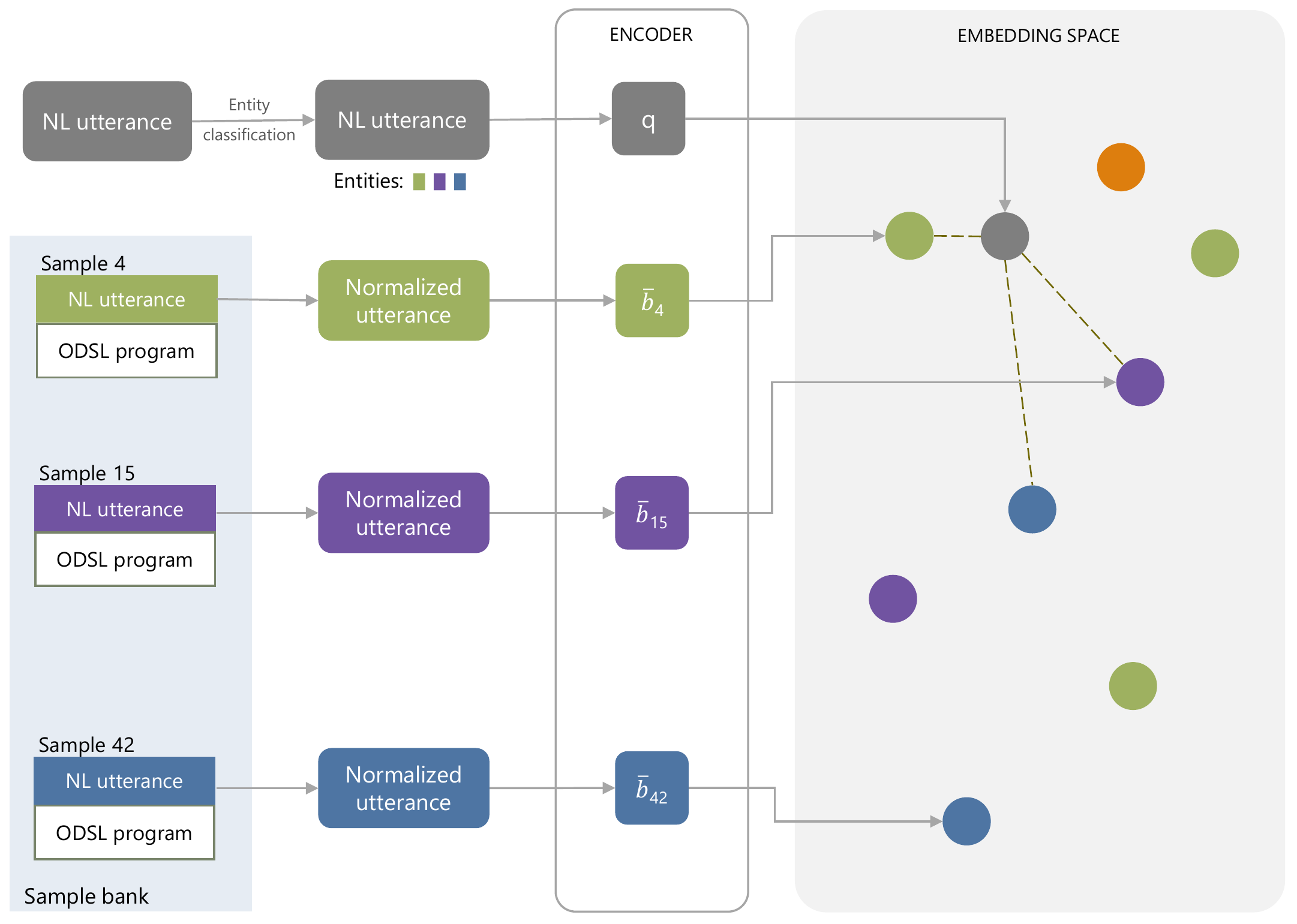}
     \caption{Illustration of the semantic search for the top 3 samples most relevant to a given NL utterance. Each entity type is encoded with a different color.}
     \label{fig:search}
\end{figure}

We select $\max{(k, \textrm{length}(\mathcal{E}))}$ samples from $\mathcal{B}$ such that each associated entities in $\mathcal{E}$ is represented at least once. The algorithm is described in Alg.\ref{algo:selection} and illustrated in Fig.~\ref{fig:search}.

\paragraph{Context-Aware Sub-Sample Selection}

\begin{figure}
     \centering
     \includegraphics[width=1.0\linewidth]{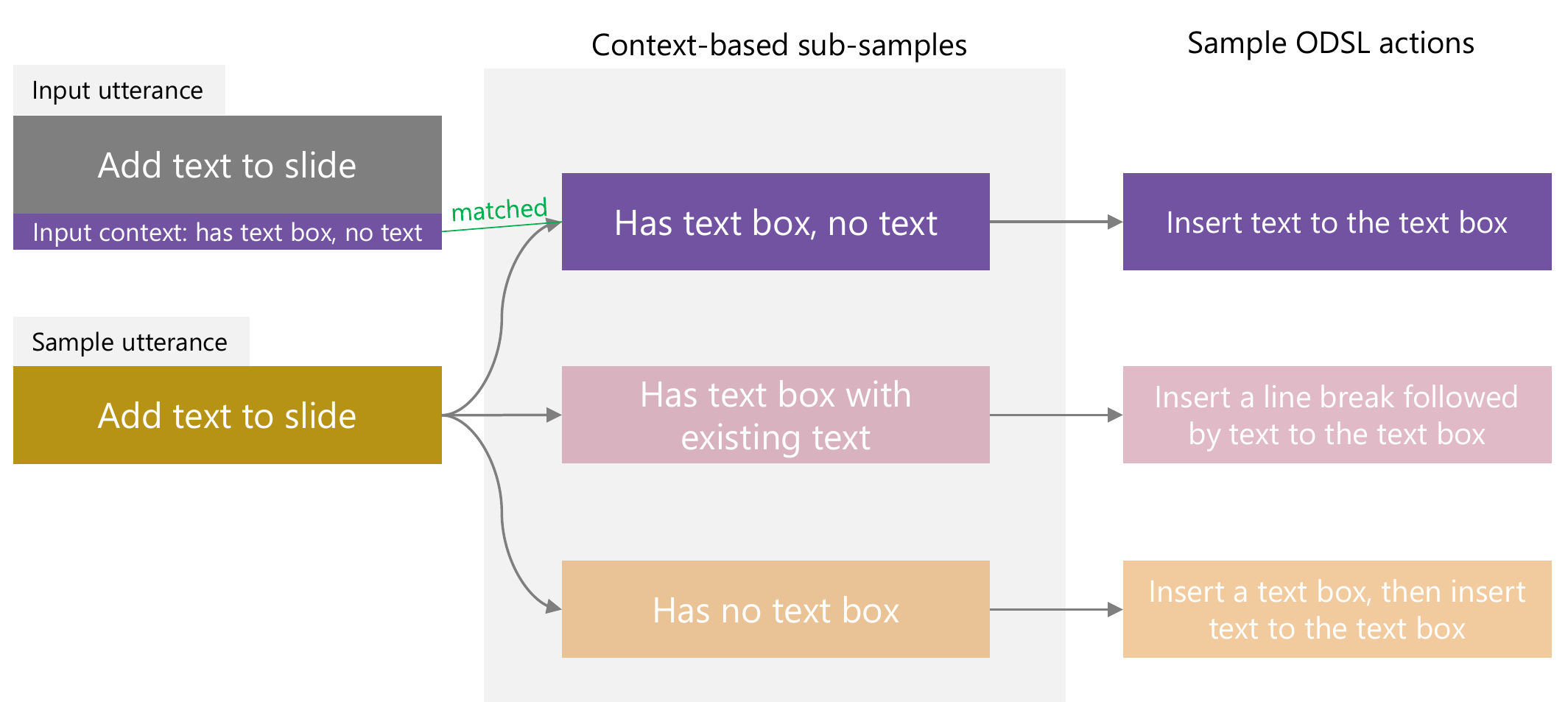}
     \caption{Example context-aware sub-sample selection process. The sample utterance \enquote{Add text to slide} has 3 sub-sample ODSL implementations corresponding to different document contexts.}
     \label{fig:subselection}
\end{figure}

\dsl samples with context can have context-based \textit{sub-samples}. These \textit{sub-samples} have different \dsl implementations for the same sample utterance based on variations in the document context $c$. For example, as illustrated in Fig. \ref{fig:subselection}, when adding a sentence to a slide, depending on the current state of the slide, we will want to (a) insert the text directly if the slide contains a blank text box, (b) insert a line break followed by the text if there is existing text on the slide, or (c) create a text field before inserting the text if the slide does not contain a text box. When retrieving programs to include in the prompt, we choose the program sub-sample that has a document context most similar to the current document context. 

\begin{figure}
\centering
\includegraphics[width=0.8\textwidth]{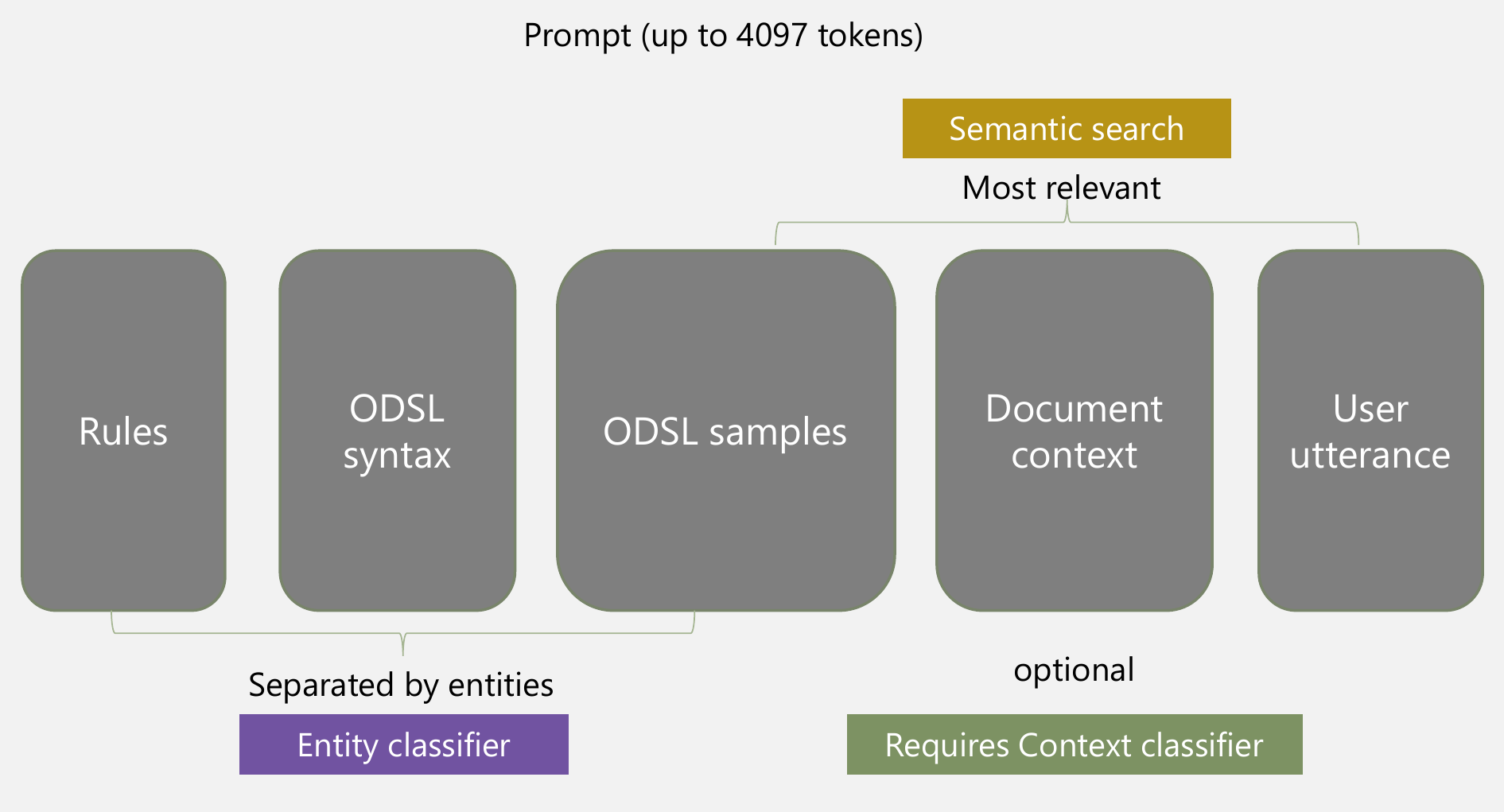}
\caption{Prompt components and their relations with Entity classifier, Requires-Context classifier, and Semantic search process. The Requires-Context classifier decides whether to include document context in the prompt. The Entity classifier chooses which rules and ODSL syntax to include. Entity and Requires-Context classifiers also condition the Semantic search process to select appropriate ODSL samples.}
\label{fig:prompt_component}
\end{figure}

Fig. \ref{fig:prompt_component} shows the relations between different components of the prompt and modules in ARM. An example ODSL synthesis prompt is shown in Fig. \ref{fig:prompt}.

\begin{figure}
\begin{AIbox}{ODSL Synthesis Prompt}
\lstset{frame=none, numbers=none, xleftmargin=0cm}
\textbf{System instruction}: ODSL is a DSL for performing actions in PowerPoint.\\
Here are examples of ODSL's syntax: 
\begin{lstlisting}
# Get the title from all slides in the presentation
textRanges = select_text(scope="Presentation", name="Title")

# Gets the textRanges matching the string "Hello" from the provided shapes.
textRanges = select_text(scope=shapes, text="Hello")  

# Formats the text in textRanges to be bold, italic, have Times New Roman font, have a single underline, have font size 24, have the color teal and be Left aligned.
format_text(textRanges=textRanges, bold=true, fontName="Times New Roman", horizontalAlignment="Left", size=24, color="teal", italic=true, underline="Single")

# Many of the argument to format statements are optional. For example, this format statement makes the text bulleted and changes its color to olive.
format_text(textRanges=textRanges, bulleted=true, color="#808000")
\end{lstlisting} 
...\\\\
 Generate an ODSL program to fulfill the given user utterance. Remember to follow the following rules when generating ODSL:
\begin{itemize}[leftmargin=5mm,label={--}]
\item For \texttt{select\_text}, if scope is provided it must be a either \texttt{Presentation} or a variable of type shapes or slides. If no scope is provided, we select the user slide selection.
\item You must select or insert an entity before formatting or deleting it.
\item Never use for loops, array indexing or if/else statements.
\\...\\
\end{itemize}
\smallskip
\textbf{User}: Change the text format to make it look like a typewriter\\
\textbf{Assistant}:
\begin{lstlisting}
text = select_text()
format_text(textRanges=text, fontName="Courier New", size=18, bold=false, italic=false, underline="None", color="#000000", bulleted=false, horizontalAlignment="Left")
\end{lstlisting}
\textbf{User}: Change the text format to make it look elegant\\
\textbf{Assistant}:
\begin{lstlisting}
text = select_text()
format_text(textRanges=text, fontName="Times New Roman", size=18, italic=true)
\end{lstlisting}
...\\ \\
\textbf{User}: Make the text look formal\\
\textbf{Assistant}:
\end{AIbox}
\caption{Example LLM prompt to synthesize an \dsl program for user utterance \enquote{Make the text look formal.} This utterance does not require document context. If document context is required, the prompt is similar except that we also include document context both for the current user utterance and the few-shot \dsl samples. The prompt shown is shortened for visualization.}
\label{fig:prompt}
\end{figure}

\section{Evaluation Procedure}\label{sec:eval}

\begin{figure}

\begin{minipage}{.5\linewidth}
\centering
\subfloat[Flowchart of evaluation procedure.]{\label{fig:evaluation:a}\includegraphics[scale=.5]{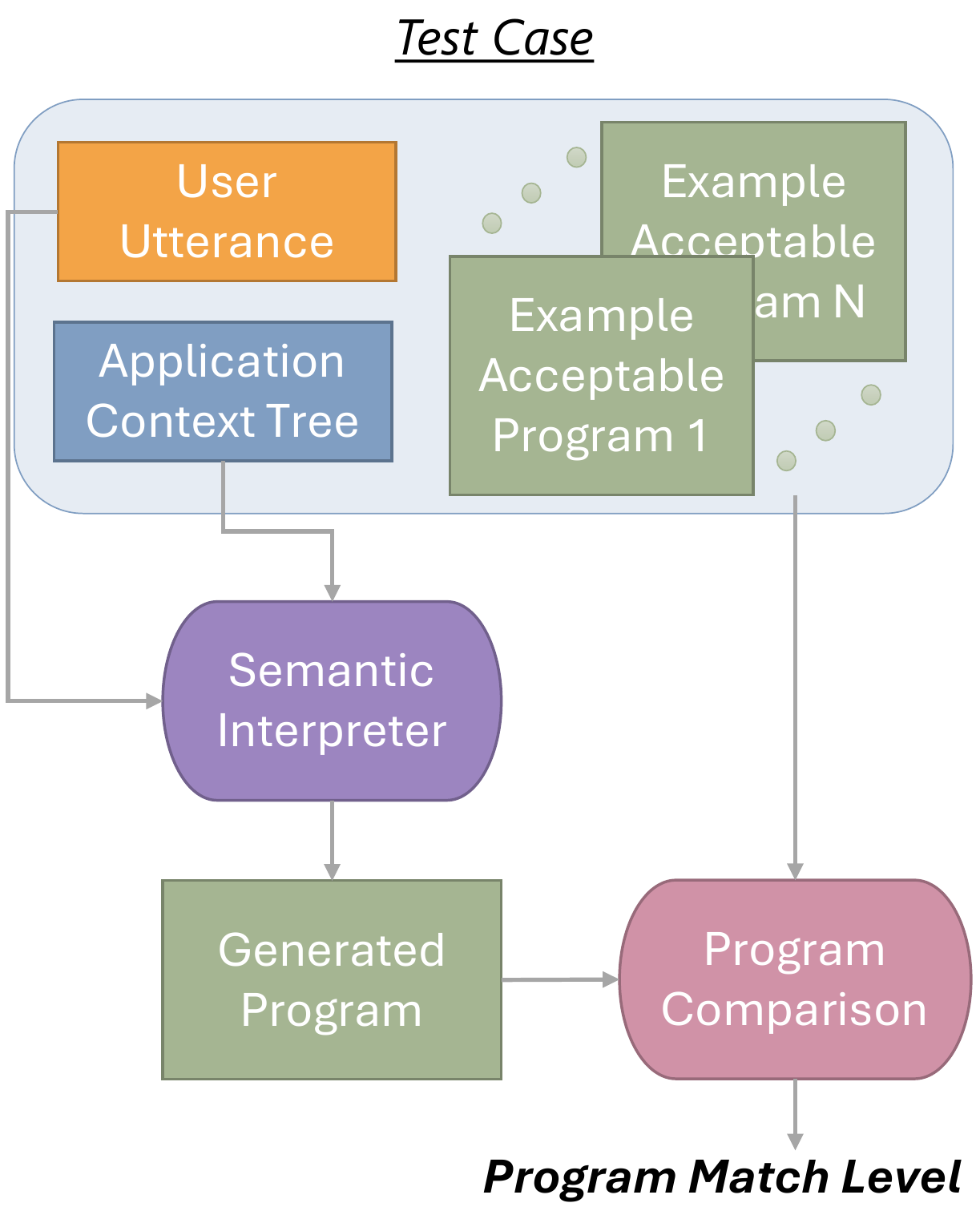}}
\end{minipage}
\begin{minipage}{.5\linewidth}
\centering
\subfloat[Program normalization performs transforms such as canonicalization and desensitization to avoid spurious discrepancies that prevent matches.]{\label{fig:evaluation:b}\includegraphics[scale=.48]{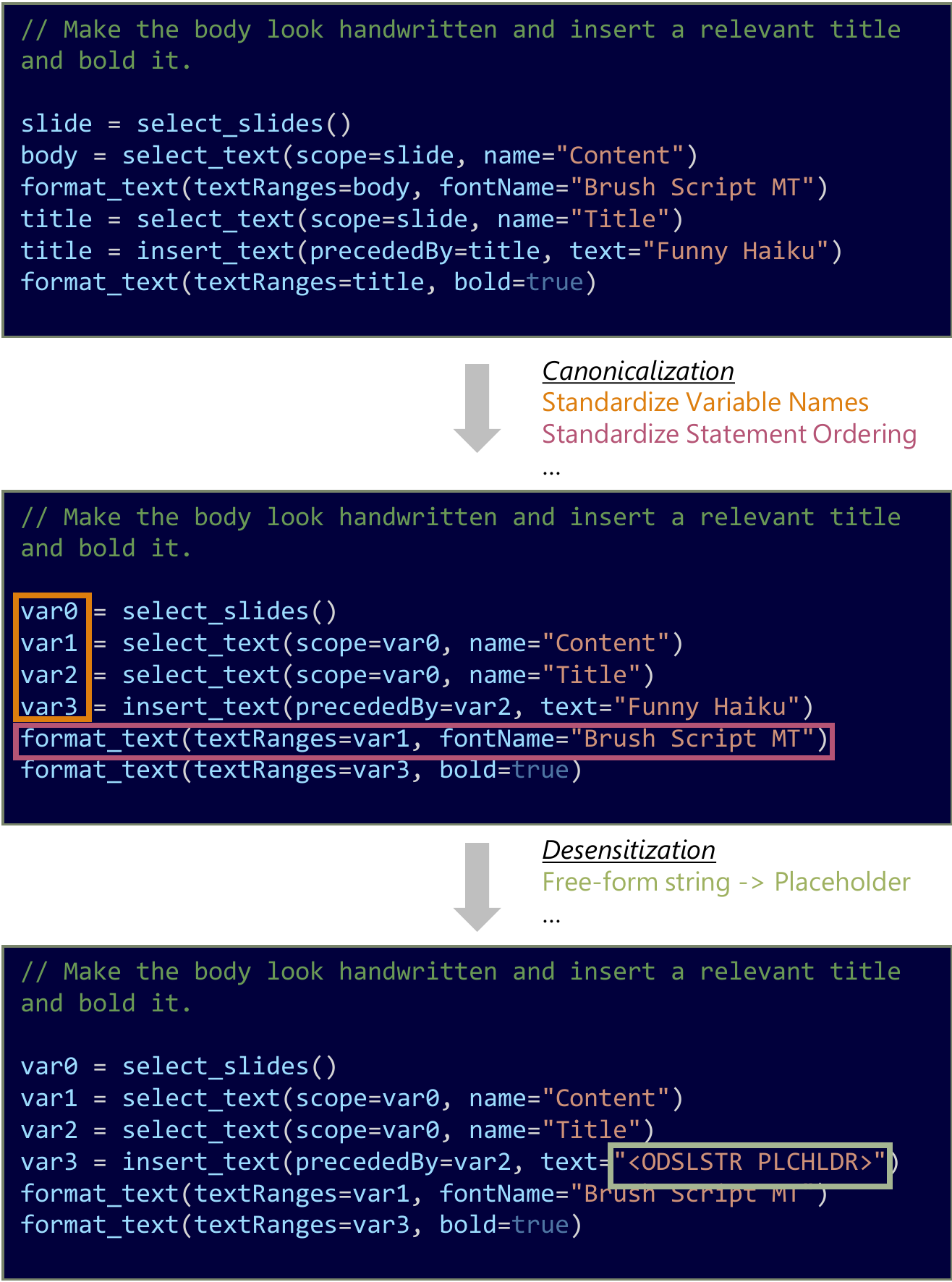}}
\end{minipage}\par\medskip
\centering
\subfloat[Subprogram analysis determines if an acceptable program is contained within the generated program.]{\label{fig:evaluation:c}\includegraphics[scale=.55]{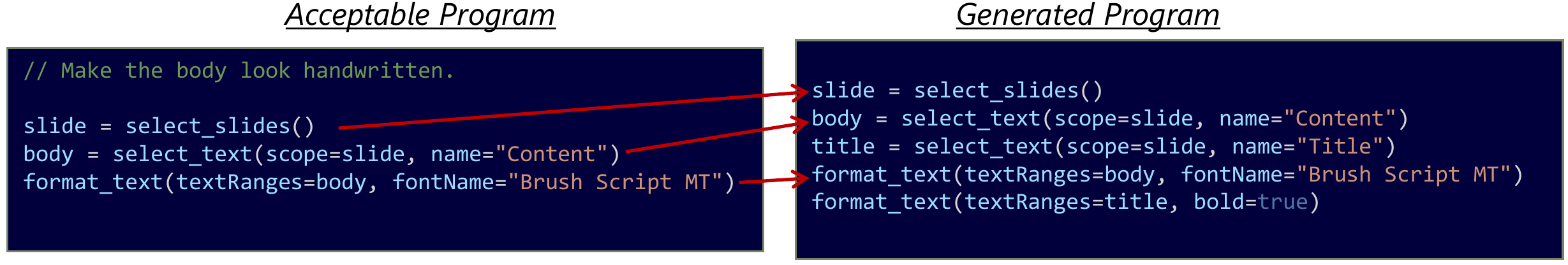}}

\caption{Evaluation Procedure using program equivalence analysis.}
\label{fig:evaluation}
\end{figure}

Evaluation of a natural language commanding system like \system is non-trivial for two main reasons: (1) the set of possible user utterances is unbounded; (2) the number of valid ways to fulfill a user utterance can be unbounded. To help address the former, we created an evaluation set with 197 tests cases with user utterances that span only scenarios that our \dsl exploration for PowerPoint is able to express today: creating presentations, adding new slides, inserting text content, modifying or rewriting existing content, inserting images, formatting entities in the document, etc. We do not include test cases for functionality that our \dsl exploration does not currently support; e.g., creating charts, file sharing, creating or resolving comments, etc. 

To understand the latter challenge of having multiple possible solutions for the same user utterance, consider the user utterance, \enquote{Make the slide look beautiful}: there is no single correct interpretation. The system may choose to change font properties, insert relevant images, change the slide layout, insert shapes, animate the slide, or some subset of these. In this work, given that \system translates a user intent to a DSL program, we propose a procedure that reformulates the problem of evaluating the natural language commanding systems into the problem of analyzing program equivalence.

\subsection{Test Case Format}
Each test case consists of the user utterance, optional document context, and a list of acceptable programs as shown in Fig.~\ref{fig:evaluation:a}. Allowing a list of acceptable programs rather than just a one allows us to include multiple known correct interpretations of the user's intent and also keeps the test case extensible as we encounter new acceptable interpretations. During evaluation we pass the user utterance to the \system to obtain the generated program. The generated program is then compared to each of the acceptable programs using a program comparison algorithm discussed below. The program comparison algorithm outputs a \textit{program match level} which indicates degree of program match. We discuss the different levels of match we use below.

\subsection{Program Comparison}
The program comparison first parses the \dsl programs into Abstract Syntax Tree (AST) representations. We then perform a series of program transformations and analyses to obtain the appropriate program match level as an output. Particularly, we use the following two ideas that we refer to as \textit{program normalization} and \textit{subprogram analysis}.

\paragraph{Program Normalization} The program normalization step involves program transformations that desensitize program comparison to spurious discrepancies. Program normalization first performs canonicalization: this transforms the program to an equivalent program that uses standardized conventions (e.g., standardizing equivalent statement orderings, standardizing variable names, etc.) Next, we perform desensitization where we can take free-form parameters like strings or numerical parameters (where appropriate) and replace them with placeholders to desensitize program comparison to their values. An example is shown in Fig.~\ref{fig:evaluation:b}.

\paragraph{Subprogram Analysis} We noticed that often a generated program may perform all the steps in an acceptable program but also perform additional unnecessary but benign steps on top of these. Fig.~\ref{fig:evaluation:c} shows an example where the generated program performs the same steps to satisfy the user utterance of \enquote{Make the body look handwritten} by changing the font name, but then also does some additional formatting on top by making the text bold. Here the generated program is still valid for the user utterance: from a user's perspective the resulting text still looks handwritten after program execution regardless of whether it was bolded or not. This case is not an exception, e.g., a generated program may insert a picture and then resize it where an acceptable program may just insert the picture; a \enquote{create a presentation about X} intent may create 3 slides in an acceptable response but 5 slides in the generated response. To evaluate such cases fairly, without having to manually add all such cases to the list of acceptable responses, we perform subprogram analysis. This analysis checks if an acceptable program is contained within a generated program, i.e., does the generated program do at least what the acceptable program performs. Note that a successful subprogram match may not always be acceptable; e.g., for a user intent about inserting a picture, a generated program could insert an image but incorrectly delete it later on. In practice, we do not observe such behavior in the generated programs for our test cases and so deem subprogram matches to be considered as a pass. 

The output of the evaluation procedure is a \textit{program match level}. Here are the different program match levels in descending order of degree of match:
\begin{itemize}
    \item \textbf{Exact} \textendash~At least one acceptable program is an exact match to the generated program (no normalization - must be a strict match).
    \item \textbf{Normalized} \textendash~ At least one acceptable program is equivalent to the generated program after program normalization.
    \item \textbf{Subprogram Exact} \textendash~At least one acceptable program is contained within the generated program (no normalization - must be strict subprogram).
    \item \textbf{Subprogram Normalized} \textendash~At least one acceptable program is contained within the generated program after program normalization.
    \item \textbf{Manual Check - Valid} \textendash~Due to the creative nature of LLMs, even if a test case result does not fall into any of the above match levels, there is still a chance that the output is correct. For this we perform manual human analysis of such test cases and check if the program is valid.
    \item \textbf{None} \textendash~\dsl program produced but the program match does not satisfy any of the above criteria.
    \item \textbf{Error} \textendash~No valid \dsl program produced due to errors (e.g., no program output, incomplete output, syntax errors, timeout errors, etc.)
\end{itemize}

In practice, we found that categories of Subprogram Normalized and above led to valid programs. Thus, we define pass rate for our experiments as the percentage of test cases that have a program match level of Subprogram Normalized or above. 

\section{Experiments}
\label{sec:exp}
We use the evaluation procedure described in Section~\ref{sec:eval} to perform an ablation study to understand the contribution of the various components in the \system. Table~\ref{table:results} shows the variants of Semantic Interpreter and the evaluation results. In the variant with $k = 0$ for top-$k$ sample selection, we provide \dsl syntax description and rules in the prompts, but we do not provide any full \dsl program examples. For other values of $k$, we follow the procedure as described in Section~\ref{subsec:retrieval}, providing $\max(k, \textrm{length}(\mathcal{E}))$ examples where $\mathcal{E}$ is the set of matched entities.

For all our experiments, we use the \texttt{text-davinci-003} model from OpenAI's GPT-3.5 model family\footnote{\url{https://platform.openai.com/docs/models}}. This model has a context length of 4097 tokens. We set \texttt{temperature} to 0 and \texttt{top-p} to 1.

\begin{table}[t!]
\resizebox{\textwidth}{!}{%
\begin{tabular}{@{}cccccccccccc@{}}
\toprule
\makecell[t]{Entity \\ Classifier} & \makecell[t]{Requires Context \\ Classifier} & \makecell[t]{Code \\ Correction} & \makecell[c]{$k$} & \makecell[t]{Exact} & \makecell[t]{Normalized} & \makecell[t]{Subprogram \\ Exact} & \makecell[t]{Subprogram \\ Normalized} & \makecell[t]{Manual Check \\ Valid} & \makecell[t]{None} & \makecell[t]{Error} & \makecell[t]{Pass Rate \\ (\%)} \\
\midrule
\cmark & \cmark & \cmark & 0 & 0 & 10 & 0 & 1 & 8 & 28 & 150 & $10.42 \pm 4.22$  \\
 \cmark & \cmark & \cmark & 1 & 20 & 89 & 5 & 40 & 20 & 13 & 10 & $87.59 \pm 4.56$ \\
 \cmark & \cmark & \cmark & 3 & 28 & 91 & 8 & 39 & 19 & 6 & 6 & $93.07 \pm 3.51$ \\
 \midrule
 \cmark & \cmark & \cmark & 5 & 37 & 96 & 9 & 37 & 12 & 1 & 5 & $\mathbf{96.06 \pm 2.69}$ \\
 \midrule
  \cmark & \cmark & \cmark & 7 & 32 & 92 & 10 & 32 & 21 & 1 & 9 & $94.06 \pm 3.27$ \\
  \cmark & \cmark & \cmark & 9 & 26 & 66 & 8 & 40 & 24 & 4 & 29 & $82.61 \pm 5.24$ \\
  \xmark & \cmark & \cmark & 5 & 17 & 89 & 8 & 38 & 16 & 15 & 14 & $84.60 \pm 4.99$ \\
  \cmark & \xmark & \cmark & 5 & 21 & 69 & 5 & 42 & 18 & 34 & 8 & $78.13 \pm 5.72$ \\
  \xmark & \xmark & \cmark & 5 & 14 & 75 & 5 & 49 & 15 & 28 & 11 & $79.62 \pm 5.57$ \\
  \cmark & \cmark & \xmark & 5 & 33 & 93 & 8 & 36 & 11 & 1 & 15 & $91.08 \pm 3.94$ \\
  \bottomrule

\end{tabular}%
}
\vspace{1mm}
\caption{Results of ablation study, running the evaluation procedure discussed in Section~\ref{sec:eval} on variants of \system. The statistical uncertainties are reported at 95\% confidence level using Agresti-Coull interval. Note that while the pass rates are quite high for certain configurations, the experiments are run on a benchmark that only includes test cases for user utterances that can be expressed in our current \dsl implementation.}
\label{table:results}
\end{table}

\begin{figure}
    \centering
    \hspace*{-1.0cm}\includegraphics[width=0.65\textwidth]{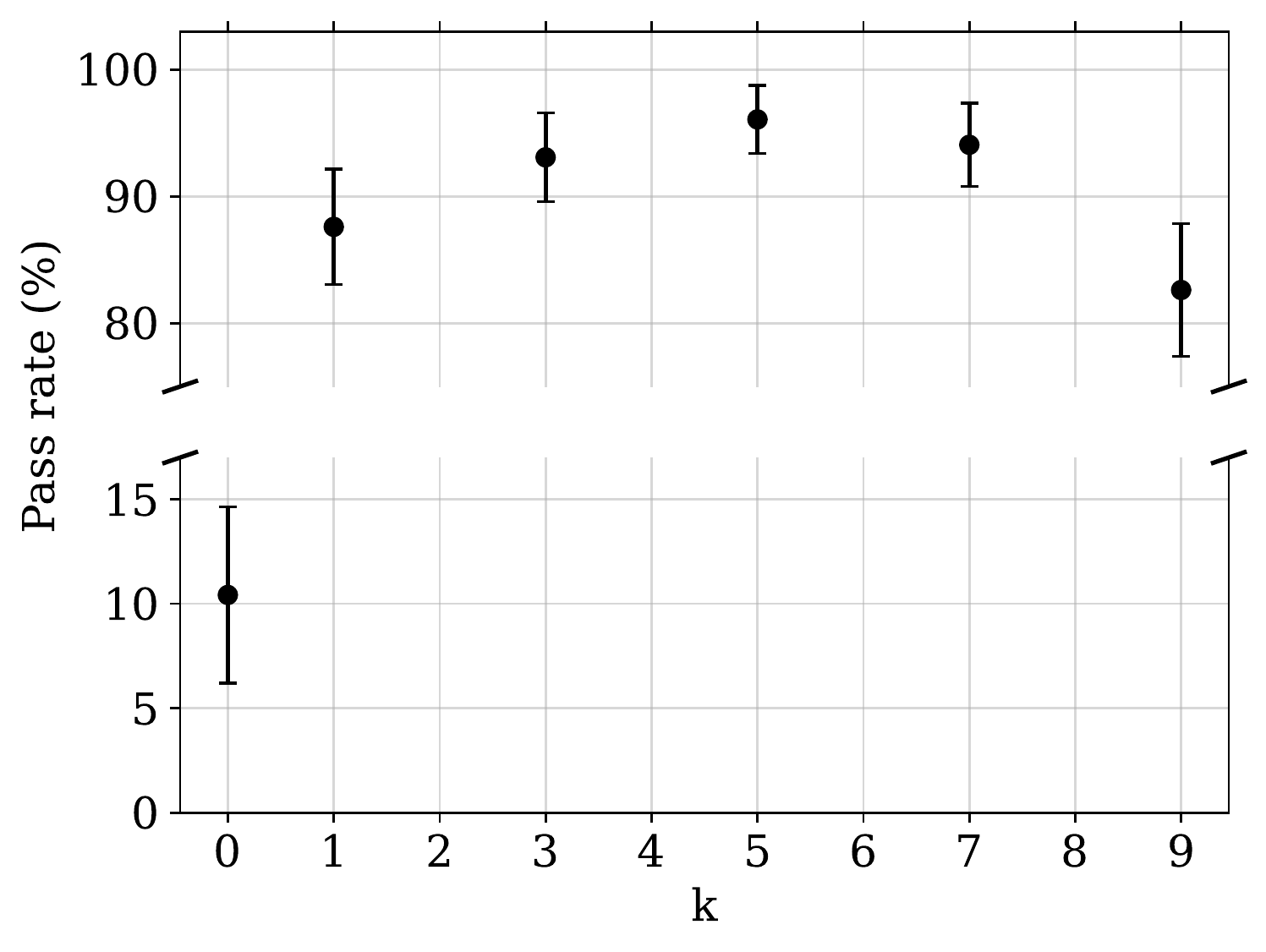}
    \caption{Visualization of pass rates for different $k$ values in ARM's retrieval as specified in the first 6 rows of Table \ref{table:results}. At $k=0$, only ODSL syntax description and rules are included in the prompt and there are no \dsl program examples. For $k>0$, $\max(k,\textrm{length}(\mathcal{E}))$ \dsl program samples are included. The error bars represent Agresti-Coull intervals at 95\% confidence level.}
    \label{fig:pass_rate}
\end{figure}

The results of the ablation study are shown in Table~\ref{table:results}. The best performing system achieves a pass rate of $96.06 \pm 2.69 \%$  by leveraging the entity classifier, requires-context classifier, code correction and configuring $k=5$ for semantic search. The remainder of this section discusses some observed trends.

\paragraph{Effect of varying \textit{k}} As illustrated in Fig.~\ref{fig:pass_rate}, few-shot prompting is critical to the performance of the system. Setting $k=0$ (no complete \dsl program examples in the prompt) leads to a very poor performance of approximately 10\% pass rate. This is much lower performance than that of the variants that have $k > 0$. Even with $k>0$, the system performance is quite sensitive to the value of $k$ with the optimal value in our experiments being $k = 5$. We see an increase in performance as we increase the value of $k$ from 1 to 5: increasing the number of example \dsl programs in the prompt gives the model more data to ground its generation in. But providing too many examples leads to issues such as not leaving enough tokens for a complete response or the LLM overfitting on the examples provided. This is consistent with the increase in Error and None cases in Table~\ref{table:results} as we increase $k$ from 5 to 9.  

\paragraph{Benefit of Analysis for Semantic Search} As part of our semantic search process, we first analyze the query with the entity classifier and requires-context classifier to narrow down the search before performing vector search. Table~\ref{table:results} shows that in variants where we remove either one or both of these analysis steps, the pass rate of the system is 10-15\% lower compared to the variant that performs both steps.

\paragraph{Benefit of Code Correction} Code correction is also a good help to \system's performance. As expected, we see an increase in Error results without it, lowering the pass rate by 5\% compared to the variant with it. Code correction is a great example of how we can benefit from neurosymbolic approaches like \system, in this case combining the robustness and precision of compiler techniques like static program analysis with the power of LLMs.

\section{Conclusion}
In this paper, we presented \system, an AI system that uses LLMs for natural language commanding in productivity software. \system translates natural language user utterances to ODSL programs, which are domain-specific symbolic representations for manipulating content and actions in Office applications. To do this, \system leverages an Analysis-Retrieval Method (ARM) for prompt construction to generate ODSL programs using LLMs. A natural language commanding module like \system can prove to be a powerful building block in architecting more general LLM-powered assistive experiences in the productivity space.  Our work illustrates the promise of using a program synthesis approach to tackle the problem of natural language commanding, enhancing how people interact with productivity software in the future.

\section*{Acknowledgements}
We are thankful for the support of many from the Office AI team at Microsoft. Special thanks to Bobby Kishore and Sumit Chauhan for championing this work; Shavinder Multani, Nick Thomson, Kevin Nickel, Mahesh Sasidharan, Dhruvil Gala, Yijia Xu, Yipeng Li, Jignesh Shah and Chandan Mishra for all the engineering support; Alan Liu, Manan Sanghi, Chris Pleasants, Firoz Shaik, Md Muksitul Haque and Weiyao Xie for their contributions to \dsl explorations across apps.

\bibliographystyle{unsrt} 
\bibliography{ref}

\appendix



\end{document}